\documentclass[11pt,a4paper]{article}
\usepackage[hyperref]{acl2020}
\usepackage{times}
\usepackage{latexsym}
\usepackage{amsfonts}
\usepackage{graphicx}
\usepackage{caption}
\usepackage{subcaption}
\usepackage{float}
\usepackage{booktabs}

\usepackage{microtype}

\usepackage{makecell}
\usepackage{tabularx}

\aclfinalcopy

\title{From Machine Reading Comprehension to Dialogue State Tracking: Bridging the Gap}

\author{Shuyang Gao$\bf{^*}$$^{,1}$~~~Sanchit Agarwal$\bf{^*}$$^{,1}$~~~Tagyoung Chung$^1$~~~Di Jin$^2$~~~Dilek Hakkani-Tur$^1$ \\
$^1$Amazon Alexa AI, Sunnyvale, CA, USA\\
$^2$Computer Science \& Artificial Intelligence Laboratory, MIT, MA, USA \\
\{shuyag,agsanchi, tagyoung, hakkanit\}@amazon.com, jindi15@mit.edu
}

\date{}

\begin{document}
\maketitle
\begin{abstract}
Dialogue state tracking (DST) is at the heart of task-oriented dialogue systems. However, the scarcity of labeled data is an obstacle to building accurate and robust state tracking systems that work across a variety of domains. Existing approaches generally require some dialogue data with state information and their ability to generalize to unknown domains is limited. In this paper, we propose using machine reading comprehension (RC) in state tracking from two perspectives: model architectures and datasets. We divide the slot types in dialogue state into \textit{categorical} or \textit{extractive} to borrow the advantages from both \textit{multiple-choice} and \textit{span-based} reading comprehension models. Our method achieves near the current state-of-the-art in joint goal accuracy on MultiWOZ $2.1$ given full training data. More importantly, by leveraging machine reading comprehension datasets, our method outperforms the existing approaches by many a large margin in few-shot scenarios when the availability of in-domain data is limited. Lastly, even \textit{without} any state tracking data, i.e., zero-shot scenario, our proposed approach achieves greater than 90\% average slot accuracy in $12$ out of $30$ slots in MultiWOZ $2.1$\@.
\end{abstract}

\section{Introduction}
\footnotetext{*Authors contributed equally.}
Building a task-oriented dialogue system that can comprehend users' requests and complete tasks on their behalf is a challenging but fascinating problem. Dialogue state tracking (DST) is at the heart of task-oriented dialogue systems. It tracks the \textit{state} of a dialogue during the conversation between a user and a system. The \textit{state} is typically defined as the \textit{(slot\_name, slot\_value)} pair that represents, given a slot, the value that the user provides or system-provided value that the user accepts.

Despite the importance of DST in task-oriented dialogues systems, few large datasets are available. To address this issue, several methods have been proposed for data collection and bootstrapping the DST system. These approaches either utilize Wizard-of-Oz setup via crowd sourcing~\citep{wen2017network, budzianowski2018multiwoz} or Machines Talking To Machines (M2M) framework~\citep{shah2018bootstrapping}. Currently the most comprehensive dataset with state annotation is MultiWOZ~\citep{budzianowski2018multiwoz}, which contains seven domains with around $10,000$ dialogues. However, compared to other NLP datasets, MultiWOZ is still relatively small, especially for training data-intensive neural models. In addition, it is also a non-trivial to get a large amount of clean labeled data given the nature of task-oriented dialogues~\citep{eric2019multiwoz}.

Another thread of approaches have tried to utilize data in a more efficient manner. These approaches~\citep{wu2019transferable, zhou2019multi} usually train the models on several domains and perform zero-shot or few-shot learning on unseen domains. However, these methods require slot definitions to be similar between the training data and the unseen test data. If such systems are given a completely new slot type, the performance would degrade significantly.
Therefore, these approaches still rely on considerable amount of DST data to cover a broad range of slot categories.

We find machine reading comprehension task (RC)~\citep{rajpurkar2016squad, chen2018neural} as a source of inspiration to tackle these challenges. The RC task aims to evaluate how well machine models can understand human language, whose goals are actually similar to DST. Ultimately, DST focuses on the contextual \textit{understanding} of users' request and inferring the \textit{state} from the conversation, whereas RC focuses on the general understanding of the text regardless of its format, which can be either passages or conversations. In addition, recent advances have shown tremendous success in RC tasks. Thus, if we could formulate the DST task as a RC task, it could benefit DST in two aspects: first, we could take advantage of the fast-growing RC research advances; second, we could make use of the abundant RC data to overcome the data scarcity issue in DST task.

Building upon this motivation, we formulate the DST task into an RC task by specially designing a question for each slot in the dialogue state, similar to~\citet{gao2019dialog}. Then, we divide the slots into two types: \textit{categorical} and \textit{extractive}, based on the number of slot values in the ontology. For instance, in MultiWOZ, slots such as \textit{parking} take values of \textit{\{Yes, No, Don't Care\}} and can thus be treated as \textit{categorical}. In contrast, slots such as \textit{hotel-name} may accept an unlimited number of possible values and these are treated as \textit{extractive}. Accordingly, we propose two machine reading comprehension models for dialogue state tracking. For categorical slots, we use multiple-choice reading comprehension models where an answer has to be chosen from a limited number of options. And for the extractive dialogue state tracking, span-based reading comprehension are applied where the answer can be found in the form of a span in the conversation.


To summarize our approach and contributions:

\begin{itemize}
    \itemsep0em
	\item We divide the dialogue state slots into categorical and extractive types and use RC techniques for state tracking. Our approach can leverage the recent advances in the field of machine reading comprehension, including both multiple-choice and span-based reading comprehension models.
	\item We propose a two-stage training strategy. We first coarse-train the state tracking models on reading comprehension datasets, then fine-tune them on the target state tracking dataset.
	\item We show the effectiveness of our method under three scenarios: First, in full data setting,
	we show our method achieves close to the current state-of-the-art on MultiWoz $2.1$ in terms of joint goal accuracy. Second, in few-shot setting, when only $1$--$10$\% of the training data is available, we show our methods
	significantly outperform the previous methods for $5$ test domains in MultiWoz $2.0$\@. In particular, we achieve $45.91$\% joint goal accuracy with just $1$\% (around 20--30 dialogues) of 
	hotel domain data as compared to previous best result of $19.73$\% \citep{wu2019transferable}. Third, in zero-shot setting where no state tracking data is used for training, our models still achieve considerable average slot accuracy. More concretely, we show that $13$ out of $30$ slots in MultiWOZ $2.1$ can achieve an average slot accuracy of greater than $90$\% without any training.

\end{itemize}

\section{Related Works}
Traditionally, dialogue state tracking methods \citep{liu2017end, mrkvsic2016neural, zhong2018global, nouri2018toward, lee2019sumbt} assume a fully-known fixed ontology for all slots where the output space of a slot is constrained by the values in the ontology. However, such approaches cannot handle previously unseen values and do not scale well for slots such as \textit{restaurant-name} that can take potentially unbounded set of values. To alleviate these issues \citep{rastogi2017scalable, goel2018flexible} generate and score slot-value candidates from the ontology, dialogue context $n$-grams, slot tagger outputs, or a combination of them. However, these approaches suffer if a reliable slot tagger is not available or if the slot value is longer than the candidate $n$-grams. \citet{xu2018end} proposed attention-based pointing mechanism to find the start and end of the slot value to better tackle the issue of unseen slot values. \citet{gao2019dialog} proposed using a RC framework for state tracking. They track slot values by answering the question ``what is the value of the slot?'' through attention-based pointing to the dialogue context. Although these approaches are more practical and scalable, they suffer when the exact slot value does not appear in the context as expected by the backend database or if the value is not \textit{pointable}. More recently, hybrid approaches have attempted to combine the benefits of both using predefined ontology (closed vocabulary) and dynamically generating candidate set or pointing (open vocabulary) approaches. \citet{goel2019hyst} select between the two approaches per slot based on dev set. \citet{wu2019transferable} utilize pointer generator network to either copy from the context or generate from vocabulary. 

Perhaps, the most similar to our work is by \citet{zhang2019find} and \citet{zhou2019multi} where they divide slot types into span-based (extractive) slots and pick-list (categorical) slots and use QA framework to point or pick values for these slots. A major limitation of these works is that they utilize heuristics to determine which slots should be categorical and which non-categorical. Moreover, in these settings most of the slots are treated as categorical (21/30 and 25/30), even though some of them have very large number of possible values, e.g., \textit{restaurant-name}. This is not scalable especially when the ontology is large, not comprehensive, or when new domains/slots can occur at test time as in DSTC8 dataset~\citep{rastogi2019towards}.

There are recent efforts into building or adapting dialog state tracking systems in low data scenarios \citet{wu2019transferable, zhou2019multi}. The general idea in these approaches is to treat all but one domain as in-domain data and test on the remaining unseen domain either directly (zero shot) or after fine-tuning on small percentage (1\%-10\%) of the unseen domain data (few shot). A major drawback of these approaches is that they require several labeled in-domain examples in order perform well on the unseen domain. This limits these approaches to in-domain slots and slot definitions and they do not generalize very well to new slots or completely unseen target domain. This also requires large amount of labeled data in the source domain, which may not be available in real-world scenario. Our proposed approach, on the other hand, utilizes domain-agnostic QA datasets with zero or a small percentage of DST data and significantly outperforms these approaches in low-resource settings.
\section{Methods}
\label{sec:method}

\begin{table*}[ht]
\centering
\small
\begin{tabular}{lcccc}
\Xhline{2\arrayrulewidth}
                   \textbf{Slot Name}         & \textbf{\# Possible Values }     & \textbf{Exact Match Rate}          &  \textbf{Extractive}          & \textbf{Categorical}         \\ \hline
hotel.semi.type & 3 & 61.1\% & $\times$  & \checkmark \\
hotel.semi.internet & 3 & 62.1\% & $\times$  & \checkmark \\
hotel.semi.parking & 4 & 63.1\% & $\times$  & \checkmark \\\hline
restaurant.semi.pricerange & 4 & 97.8\% & \checkmark  & \checkmark\\
hotel.semi.pricerange & 6 & 97.7\% & \checkmark  & \checkmark \\
hotel.semi.area & 6 & 98.8\% & \checkmark  & \checkmark \\
attraction.semi.area & 6 & 99.0\% & \checkmark  & \checkmark \\
restaurant.semi.area & 6 & 99.2\% & \checkmark  & \checkmark \\
hotel.semi.stars & 7 & 99.2\% & \checkmark  & \checkmark \\
hotel.book.people & 8 & 98.2\% & \checkmark  & \checkmark \\
hotel.book.stay & 8 & 98.9\% & \checkmark  & \checkmark \\
train.semi.day & 8 & 99.3\% & \checkmark  & \checkmark \\
restaurant.book.day & 8 & 98.7\% & \checkmark  & \checkmark \\
restaurant.book.people & 8 & 99.1\% & \checkmark  & \checkmark \\
hotel.book.day & 11 & 98.1\% & \checkmark  & \checkmark \\ \hline
train.book.people & 12 & 94.7\%  & \checkmark  & $\times$ \\  
train.semi.destination & 27 & 98.2\% & \checkmark  & $\times$\\
attraction.semi.type & 27 & 86.6\% & \checkmark  & $\times$\\
train.semi.departure & 31 & 97.6\% & \checkmark  & $\times$\\
restaurant.book.time & 67 & 97.2\% & \checkmark  & $\times$\\
hotel.semi.name & 78 & 88.7\% & \checkmark  & $\times$\\
taxi.semi.arriveby & 97 & 91.9\% & \checkmark  & $\times$\\
restaurant.semi.food & 103 & 96.4\% & \checkmark  & $\times$\\
taxi.semi.leaveat & 108 & 81.1\% & \checkmark  & $\times$\\
train.semi.arriveby & 156 & 91.5\% & \checkmark  & $\times$\\
attraction.semi.name & 158 & 84.3\% & \checkmark  & $\times$\\
restaurant.semi.name & 182 & 93.9\% & \checkmark  & $\times$\\
train.semi.leaveat & 201 & 87.4\% & \checkmark  & $\times$\\
taxi.semi.destination & 251 & 87.9\% & \checkmark  & $\times$\\
taxi.semi.departure & 253 & 84.6\% & \checkmark  & $\times$\\
\Xhline{2\arrayrulewidth}
\end{tabular}
\caption{\small Slot statistics for MultiWOZ $2.1$\@. We classify the slots into extractive or categorical based on their exact match rate in conversation as well as number of possible values. $3$ slots are categorical only, $12$ slots are both extractive and categorical, the remaining $15$ slots are extractive only.}
\label{table:exact_match_rate}
\end{table*}

\subsection{Dialogue State Tracking as Reading Comprehension}

\paragraph{Dialogue as Paragraph} For a given dialogue at turn $t$, let us denote the user utterance tokens and the agent utterance tokens as $\mathbf{u}_t$ and ${\mathbf{a}_t}$ respectively. We concatenate the user utterance tokens and the agent utterance tokens at each turn to construct a sequence of tokens as $\mathbf{D}_t = \{\mathbf{u}_1, \mathbf{a}_1,...,\mathbf{u}_t\}$. $\mathbf{D}_t$ can be viewed as the paragraph that we are going to ask questions on at turn $t$. 

\paragraph{Slot as Question} We can formulate a natural language question $\mathbf{q}_i$, for each slot $s_i$ in the dialogue state. Such a question describes the meaning of that slot in the dialogue state. Examples of (slot, question) pairs can be seen in Table~\ref{table:extractive_dialogue} and~\ref{table:cat_dialogue}. We formulate questions by considering characteristics of domain and slot. In this way, DST becomes finding answers $\mathbf{a}_i$ to the question $\mathbf{q}_i$ given the paragraph $\mathbf{D}_t$. Note that~\citet{gao2019dialog} formulate dialogue state tracking problem in a similar way but their question formulation \textit{``what is the value of a slot ?''} is more abstract, whereas our questions are more concrete and meaningful to the dialogue.

\subsection{Span-based RC To Extractive DST}
\begin{figure}[htbp]
    \centering
    \includegraphics[width=0.47\textwidth]{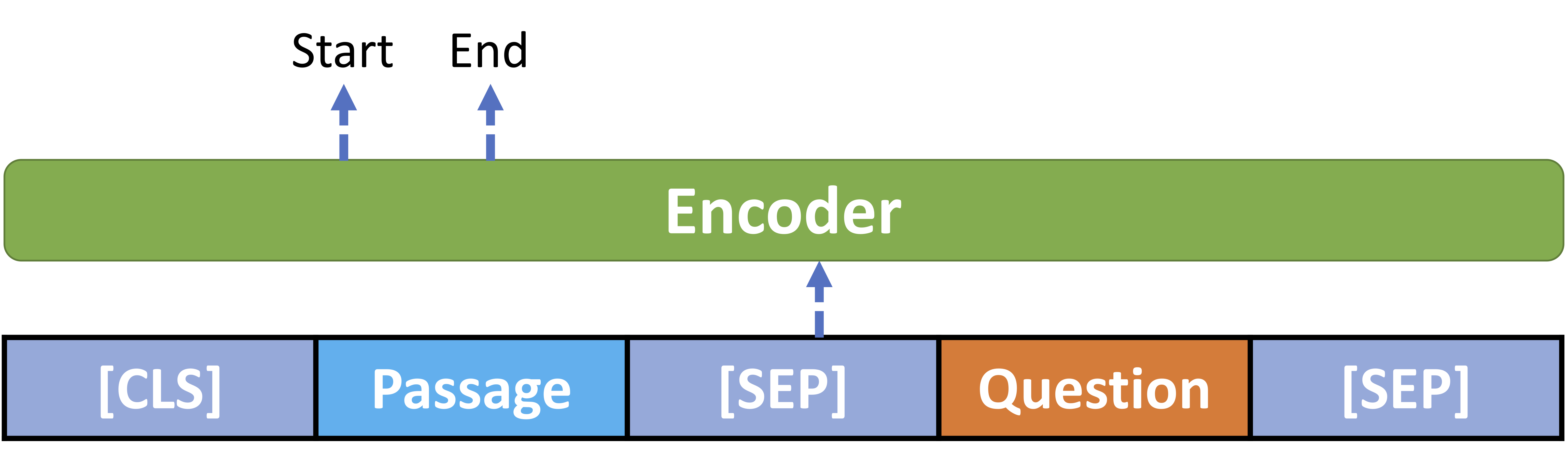}
    \caption{\small Model architecture for extractive state tracking. ``Encoder''is a pre-trained sentence encoder such as BERT.}
    \label{fig:noncat_model}
    \vspace{-2mm}
\end{figure}
\begin{table}[th!]
\centering
\small
\renewcommand{\arraystretch}{1.2}
\begin{tabularx}{\linewidth}{X}
\hline 
\textbf{Dialogue} \\ \hline
U: I'm so hungry. Can you find me a place to eat in the city centre? \\
A: I'm happy to help! There are a great deal of restaurants there. What type of food did you have in mind? \\
U: I do not care, it just needs to be expensive.\\
A: Fitzbillies restaurant serves British food would that be okay?\\
U: Yes, may I have the address?\\
\hline
\vspace{.1cm}
\textbf{restaurant.semi.food}: What type of food does the user want to eat?\\
\textbf{Answer}: \lbrack\hspace{.05cm}52-53\hspace{.05cm}\rbrack \hspace{.2cm}\textit{(I do not care, it just needs to be expensive)} \\
\hline
\vspace{.1cm}
\textbf{restaurant.semi.name}: What is the name of the restaurant where the user wants to eat?\\
\textbf{Answer}: \lbrack\hspace{.05cm}53-55\hspace{.05cm}\rbrack \hspace{.2cm}  \textit{(Fitzbillies restaurant)} \\
\hline

\end{tabularx}
\caption{\small Sample dialogue from MultiWOZ dataset showing framing of extractive DST to span-based RC. The span text (or \textit{don't care} user utterance) is also shown in italics.}
\label{table:extractive_dialogue}
\end{table}

For many slots in the dialogue state such as names of attractions, restaurants, and departure times, one can often find their values in the dialogue context with exact matches. Slots with a wide range of values fits this description. Table~\ref{table:exact_match_rate} shows the exact match rate for each slot in MultiWOZ $2.1$ dataset~\citep{budzianowski2018multiwoz, eric2019multiwoz} where slots with large number of possible values tend to have higher exact match rate ($\ge 80\%$). We call tracking such slots as \textit{extractive dialogue stack tracking (EDST)}.

This problem is similar to span-based RC where the goal is to find a span in the passage that best answers the question. Therefore, for EDST, we adopt the simple BERT-based question answering model used by \citet{devlin2019bert}, which has shown strong performance on multiple datasets~\citep{rajpurkar2016squad, rajpurkar2018know, reddy2019coqa}. 
In this model as shown in Figure~\ref{fig:noncat_model}, the slot question and the dialogue are represented as a single sequence. The probability of a dialogue token $t_i$ being the start of the slot value span is computed as $p_i = \frac{e^{\mathbf{s} \cdot \mathbf{T}_i}}{\sum _j e^{\mathbf{s} \cdot \mathbf{T}_j} }$, where $\mathbf{T}_j$ is the embedding of each token $t_j$ and $\mathbf{s}$ is a learnable vector. A similar formula is applied for finding the end of the span. 

\paragraph{Handling \textit{None} Values} At any given turn in the conversation, there are typically, many slots that have not been mentioned or accepted yet by the user. All these slots must be assigned a \textit{None} value in the dialogue state. We can view such cases as \textit{no answer exists} in reading comprehension formulation. Similar to \citet{devlin2019bert} for SQuAD $2.0$ task, we assign the answer span with start and end at the beginning token [CLS] for these slots.

\paragraph{Handling \textit{Don't Care} Values} To handle \textit{don't care} value in EDST, a span is also assigned to \textit{don't care} in the dialogue. We find the dialogue turn when the slot value first becomes \textit{don't care} and set the start and end of \textit{don't care} span to be the start and end of the user utterance of this turn. See Table~\ref{table:extractive_dialogue} for an example.
 
\subsection{Multiple-Choice Reading Comprehension to Categorical Dialogue State Tracking}
\begin{table}[h]
\centering
\small
\renewcommand{\arraystretch}{1.2}
\begin{tabularx}{\linewidth}{X}
\hline 
\textbf{Dialogue} \\ \hline
U: I am looking for a place to to stay that has cheap price range it should be in a type of hotel \\
A: Okay , Do you have a specific area you want to stay in? \\
U: No, I just need to make sure it's cheap. Oh, and I need parking.\\
\hline
\vspace{.1cm}

\textbf{hotel.semi.area}: What is the area that the user wants to book a hotel in?\\
\textbf{A. }East \hspace{.2cm} \textbf{B. } West \hspace{.2cm} \textbf{C. }North \hspace{.2cm} \textbf{D. }South \hspace{.2cm} \textbf{E. }Centre \hspace{.3cm} \textbf{F. }Don't Care \checkmark \hspace{.2cm} \textbf{G. }Not Mentioned \hspace{.2cm}\\
\hline
\vspace{.1cm}

\textbf{hotel.semi.parking}: Does the user want parking at the hotel?\\
\textbf{A. }Yes \checkmark \hspace{.2cm} \textbf{B. } No \hspace{.2cm} \textbf{C. }Don't Care \hspace{.2cm} \textbf{D. }Not Mentioned\\ \hline

\end{tabularx}
\caption{\small Sample dialogue from MultiWOZ dataset showing framing of categorical DST to multiple-choice RC.}
\label{table:cat_dialogue}
\end{table}
\begin{figure}[h]
    \centering
    \includegraphics[width=0.5\textwidth]{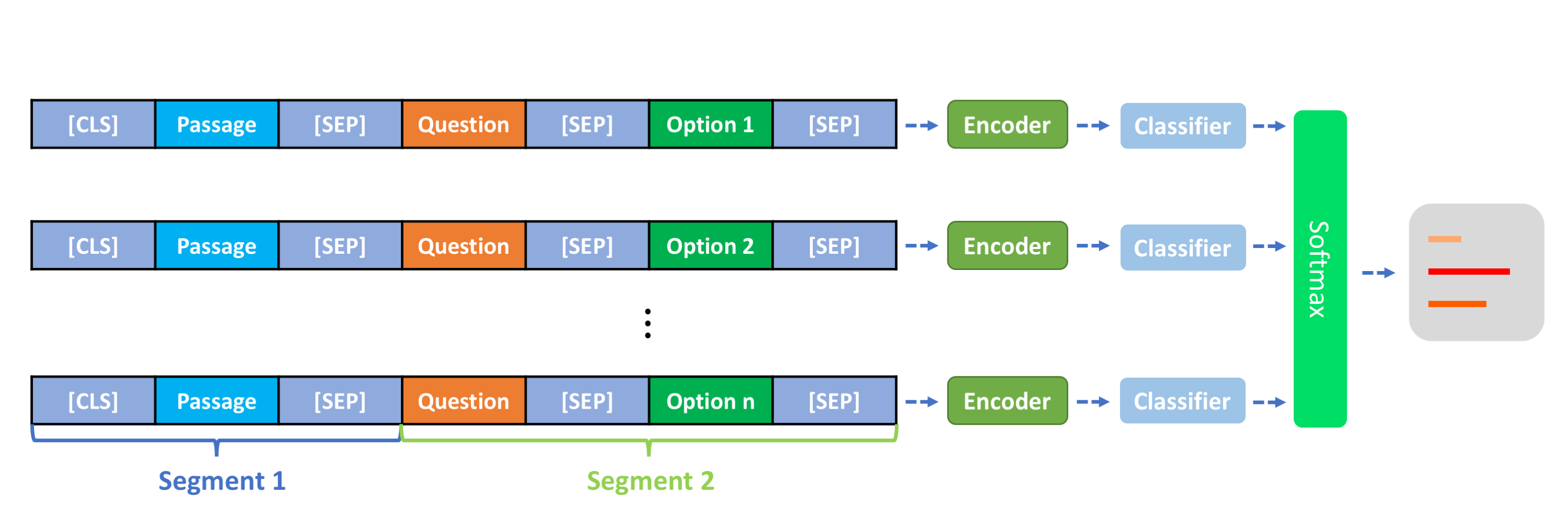}
    \caption{\small Model architecture for categorical dialog state tracking. ``Encoder''is a pre-trained sentence encoder such as BERT. ``Classifier'' is a top-level fully connected layer.}
    \label{fig:cat_model}
\end{figure}

The other type of slots in the dialogue state cannot be filled through exact match in the dialogue context in a large number of cases. For example, a user might express intent for hotel parking as \textit{``oh! and make sure it has parking''} but the slot \textit{hotel-parking} only accepts values from \textit{\{Yes, No, Don't Care\}}. In this case, the state tracker needs to infer whether or not the user wants parking based on the user utterance and select the correct value from the list. These kind of slots may not have exact-match spans in the dialogue context but usually 
require a limited number of values to choose from. 

Tracking these type of slots is surprisingly similar to multiple-choice reading comprehension (MCRC) tasks. In comparison to span-based RC tasks, the answers of MCRC datasets~\citep{lai2017race, sun2019dream} are often in the form of open, natural language sentences and are not restricted to spans in text. Following the traditional models of MCRC~\citep{devlin2019bert, jin2019mmm}, we concatenate the slot question, the dialogue context and one of the answer choices into a long sequence. We then feed this sequence into a sentence encoder to obtain a logit vector. Given a question, we can get $m$ logit vectors assuming there are $m$ answer choices. We then transform these 
$m$ logit vectors into a probability vector through a fully connected layer and a softmax layer, see Figure~\ref{fig:cat_model} for details. 

\paragraph{Handling \textit{None} and \textit{Don't Care} Values} For each question, we simply add two additional choices ``not mentioned'' and ``do not care'' in the answer options, representing \textit{None} and \textit{don't care}, as shown in Table~\ref{table:cat_dialogue}. It is worth noting that certain slots not only accept a limited number of values but also their values can be found as an exact-match span in the dialogue context. For these slots, both extractive and categorical DST models can be applied as shown in Table~\ref{table:exact_match_rate}.

\section{Experiments}
\label{sec:exp}

\subsection{Datasets}

\begin{table}[h!]
\centering
\small
\begin{tabular}{lrr}
\toprule
 &  \textbf{\# of passages} & \textbf{\# of examples}         \\ \midrule
\textbf{MRQA} (span-based) & 386,384 & 516,819     \\ 
\textbf{DREAM} (multi-choice) & 6,444 & 10,197 \\ 
\textbf{RACE} (multi-choice) & 27,933 & 97,687 \\ 
\textbf{MultiWOZ} & 8,420 &  298,978\textsuperscript{*} \\ 
\bottomrule
\end{tabular}
\caption{\small Statistics of datasets used. (*: we only report the number of positive examples (a non-empty value) in MultiWOZ for fair comparison.) }
\label{table:dataset}
\end{table}

\paragraph{MultiWOZ} We use the largest available multi-domain dialogue dataset with state annotation: MultiWOZ $2.0$~\citep{budzianowski2018multiwoz} and MultiWOZ $2.1$~\citep{eric2019multiwoz}, an enhanced, less noisier version of MultiWOZ 2.0 dataset, which contains $7$ distinct domains across 10K dialogues. We exclude \textit{hospital} and \textit{police} domain that have very few dialogues. This results in $5$ remaining domains \textit{attraction}, \textit{restaurant}, \textit{taxi}, \textit{train}, \textit{hotel} with a total of $30$ (domain, slot) pairs in the dialog state following~\citet{wu2019transferable, zhang2019find}.

\paragraph{Reading Comprehension Datasets} For span-based RC dataset, we use the dataset from Machine Reading for Question Answering (MRQA) 2019 shared task~\citep{fisch2019mrqa} that was focused on extractive question answering. MRQA contains six distinct datasets across different domains: SQuAD, NewsQA, TriviaQA, SearchQA, HotpotQA, and NaturalQuestions. In this dataset, any answer to a question is a segment of text or span in a given document. For multiple-choice RC dataset, we leverage the current largest multiple-choice QA dataset, RACE~\citep{lai2017race} as well as a dialogue-based multiple-choice QA dataset, DREAM~\citep{sun2019dream}. Both of these datasets are collected from English language exams that are carefully designed by educational experts to assess the comprehension level of English learners. Table~\ref{table:dataset} summarizes the statistics of datasets. It is worth noting that for MultiWOZ, although the number of examples are significantly more than multiple-choice QA datasets, the number of distinct questions are only 30 due to limited number of slot types.

\subsection{Canonicalization for Extractive Dialogue State Tracking}

For extractive dialogue state tracking, it is common that the model will choose a span that is either a super-set of the correct reference or has a similar meaning as the correct value but with a different wording. Following this observation, we adopt a simple canonicalization procedure after our span-based model prediction. If the predicted value does not exist in the ontology of the slot, then we match the prediction with the value in the ontology that is closest to the predicted value in terms of edit distance\footnote{we use the function \textit{get\_closest\_matches} of \textit{difflib} in Python for this implementation.}. Note that this procedure is only applied at model \textit{inference} time. At training time for extractive dialogue state tracking, the ontology is not required.

\subsection{Two-stage Training}

A two-stage training procedure is used to train the extractive and categorical dialogue state tracking models with both types of reading comprehension datasets (DREAM, RACE, and MRQA) and the dialogue state tracking dataset (MultiWOZ).

\paragraph{Reading Comprehension Training Stage} For categorical dialogue state tracking model, we coarse-tune the model on DREAM and RACE. For extractive dialogue state tracking model, we coarse-tune the model on MRQA dataset as a first step.

\paragraph{Dialog State Tracking Training Stage} After being trained on the reading comprehension datasets, we expect our models to be capable of answering (passage, question) pairs. In this phase, we further fine-tune these models on the MultiWOZ dataset.

\section{Results and Analyses}

\subsection{DST with Full Training Data}
\begin{table}[ht]
\centering
\small
\renewcommand{\arraystretch}{1.2}
\begin{tabular}{l|r}
\hline
\multicolumn{2}{c}{\textbf{Joint Goal Accuracy}} \\ \hline
SpanPtr \cite{xu2018end} & 29.09\% \\ \hline
FJST \cite{eric2019multiwoz} & 38.00\% \\ \hline
HyST \cite{goel2019hyst} & 39.10\% \\ \hline
DSTreader \cite{gao2019dialog} & 36.40\% \\ \hline
TRADE \cite{wu2019transferable} & 45.96\% \\ \hline
DS-DST \cite{zhang2019find} & \textbf{51.21}\% \\ \hline
DSTQA w/span \cite{zhou2019multi} & 49.67\%\\ \hline
DSTQA w/o span \cite{zhou2019multi} & 51.17\% \\ \hline
\textbf{STARC (this work)} & 49.48\% \\
\hline
\end{tabular}
\caption{\small Joint Goal Accuracy on MultiWOZ 2.1 test set.}
\label{table:joint_goal_acc}
\end{table}

We use the full data in MultiWOZ 2.1 to test our models. For the first $15$ slots with lowest number of possible values (from \textit{hotel.semi.type} to \textit{hotel.book.day} in Table~\ref{table:exact_match_rate}, we use our proposed categorical dialogue state tracking model whereas for the remaining 15 slots, we use the extractive dialogue state tracking model. We use the pre-trained word embedding RoBERTa-Large~\citep{liu2019roberta} in our experiment. 

Table~\ref{table:joint_goal_acc} summarizes the results. We can see that our model, STARC (\textbf{S}tate \textbf{T}racking \textbf{A}s \textbf{R}eading \textbf{C}omprehension), achieves close to the state-of-the-art accuracy on MultiWOZ 2.1 in the full data setting.
It is worth noting that the best performing approach DS-DST~\cite{zhang2019find}, cherry-picks 9 slots as span-based slots whereas the remaining 21 slots are treated as categorical. Further, the second best result DSTQA w/o span~\cite{zhou2019multi} does not use span-based model for any slot. Unlike these state-of-the-art methods, our method simply categorizes the slots based on the number of values in the ontology. As a result, our approach uses less number of (15 as compared to 21 in DS-DST) and more reasonable (only those with few values in the ontology) categorical slots. Thus, our approach is more practical to be applied in a real-world scenario. 

\begin{table}[th]
\centering
\small
\renewcommand{\arraystretch}{1.2}
\begin{tabular}{l|c}
\hline
Ablation & Dev Accuracy \\ \hline
\textbf{STARC (this work)} & \textbf{53.95}\% \\
 -- RC Coarse Tuning & 52.35\%\\
 -- Canonicalization & 51.07\% \\
 -- RC Coarse Tuning -- Canonicalization & 50.84\%\\ 
 -- Categorical Model & 47.86\% \\
 -- Categorical Model -- Canonicalization & 41.86\% \\
 \hline
DS-DST Threshold-10 & 49.08\%\\
DS-DST Span Only & 40.39\%
\\ \hline
\end{tabular}
\caption{\small Ablation study with different aspects of our model and other comparable approaches. The numbers reported are joint goal accuracy on MultiWOZ 2.1 development set.}
\label{table:ablation_studies}
\vspace{-3mm}
\end{table}

\paragraph{Ablation Study} We also run ablation study to understand which component of our model helps with accuracy. Table~\ref{table:ablation_studies} summarizes the results. For fair comparison, we also report the numbers for DS-DST Threshold-10~\cite{zhang2019find} where they also use the first 15 slots for categorical model and the remaining for extractive model. We observe that both two-stage training strategy using reading comprehension data and canonicalization play important role in higher accuracy. Without the categorical model (using extractive model for all slots), STARC is still able to achieve joint goal accuracy of 47.86\%. More interestingly, if we remove the categorical model as well as the canonicalization, the performance drops drastically, but is still slight better than purely extractive model of DS-DST.

\paragraph{Handling \textit{None} Value}

\begin{table}[ht]
\centering
\small
\scalebox{0.95}{
\begin{tabular}{l|r|r}
\hline
Error Type & Extractive & Categorical \\ \hline
ref not none, predicted none & 43.7\% & 31.4\% \\
ref none, predicted not none & 25.6\% & 58.4\% \\
ref not none, predicted not none & 30.6\% & 10.0\% \\
 \hline
\end{tabular}}
\caption{\small Type of errors made by each model.}
\label{tab:errors}
\end{table}

Through error analysis of our models, we have learned that models' performance on \textit{None} value has a significant impact on the overall accuracy. Table~\ref{tab:errors} summarizes our findings. We found that plurality errors for extractive model comes from cases where ground-truth is not \textit{None} but model predicted \textit{None}. For categorical model, the opposite was true. The majority errors were from model predicting not \textit{None} value but the ground-truth is actually \textit{None}. We leave further investigation on this issue as a future work.

\subsection{Few shot from RC to DST}

\renewcommand{\arraystretch}{1.4}
\begin{table*}[!ht]
\centering
\small
\scalebox{0.85}{
\begin{tabular}{r|r|r|r|r|r|r|r|r|r|r|r|r|r|r|r}
\hline
& \multicolumn{3}{c|}{\textbf{Hotel}}              & \multicolumn{3}{c|}{\textbf{Restaurant}}         & \multicolumn{3}{c|}{\textbf{Attraction}}       & \multicolumn{3}{c|}{\textbf{Train}}              & \multicolumn{3}{c}{\textbf{Taxi}}               \\ \cline{2-16} 
                        & \textit{1\%}   & \textit{5\%}   & \textit{10\%}  & \textit{1\%}   & \textit{5\%}   & \textit{10\%}  & \textit{1\%} & \textit{5\%}   & \textit{10\%}  & \textit{1\%}   & \textit{5\%}   & \textit{10\%}  & \textit{1\%}   & \textit{5\%}   & \textit{10\%}  \\ \hline
\textit{\textbf{TRADE}} & 19.73          & 37.45          & 41.42          & 42.42          & 55.70          & 60.94          & 35.88        & 57.55          & 63.12          & 59.83          & 69.27          & 71.11          & 63.81          & 66.58          & 70.19          \\ \hline
\textit{\textbf{DSTQA}} & N/A            & 50.18          & 53.68          & N/A            & 58.95          & 64.51          & N/A          & \textbf{70.47} & \textbf{71.60} & N/A            & 70.35          & 74.50          & N/A            & 70.90          & 74.19          \\ \hline
\textit{\textbf{STARC}}  & \textbf{45.91} & \textbf{52.59} & \textbf{57.37} & \textbf{51.65} & \textbf{60.49} & \textbf{64.66} & \textbf{40.39}        & 65.34          & 66.27          & \textbf{65.67} & \textbf{74.11} & \textbf{75.08} & \textbf{72.58} & \textbf{75.35} & \textbf{79.61} \\ 
\hline
\end{tabular}

}
\caption{\small Joint goal accuracy for few-shot experiments. Best numbers reported by TRADE and DSTQA are also shown.}
\label{table:few_shot}
\end{table*}

In few-shot setting, our model (both extractive and categorical) is pre-trained on reading comprehension datasets and we randomly select limited amount of \textit{target} domain data for fine-tuning. We do not use out-of-domain MultiWOZ data for training for few-shot experiments unlike previous works. We evaluate our model with 1\%, 5\% and 10\% of training data in the target domain. Table~\ref{table:few_shot} shows the results of our model under this setting for five domains in MultiWOZ $2.0$\@\footnote{We are showing results on MultiWOZ $2.0$ rather than $2.1$ for the purpose of comparison to previous works.}. We also report the few-shot results for other two models: TRADE~\citep{wu2019transferable} and DSTQA~\citep{zhou2019multi}, where they perform the same few-shot experiments but pre-trained with a hold-out strategy, i.e., training on the other four domains in MultiWOZ and fine-tune on the held-out domain. 
We can see that under all three different data settings, our model outperforms the TRADE and DSTQA models (expect the attraction domain for DSTQA) by a large margin. Especially in 1\% data setting for hotel domain, which contains the most number of slots (10) among all the five domains, the joint goal accuracy dropped to 19.73\% for TRADE while our model can still achieve relatively high joint goal accuracy of 45.91\%. This significant performance difference can be attributed to pre-training our models on reading comprehension datasets, which gives our model ability to comprehend passages or dialogues (which we have empirically verified in next section). The formulation of dialogue state tracking as a reading comprehension task helps the model to transfer comprehension capability. 
\subsection{Zero shot from RC to DST}
\begin{figure*}[ht!]
\begin{subfigure}[b]{0.5\linewidth}
\centering
\includegraphics[width=1\textwidth]{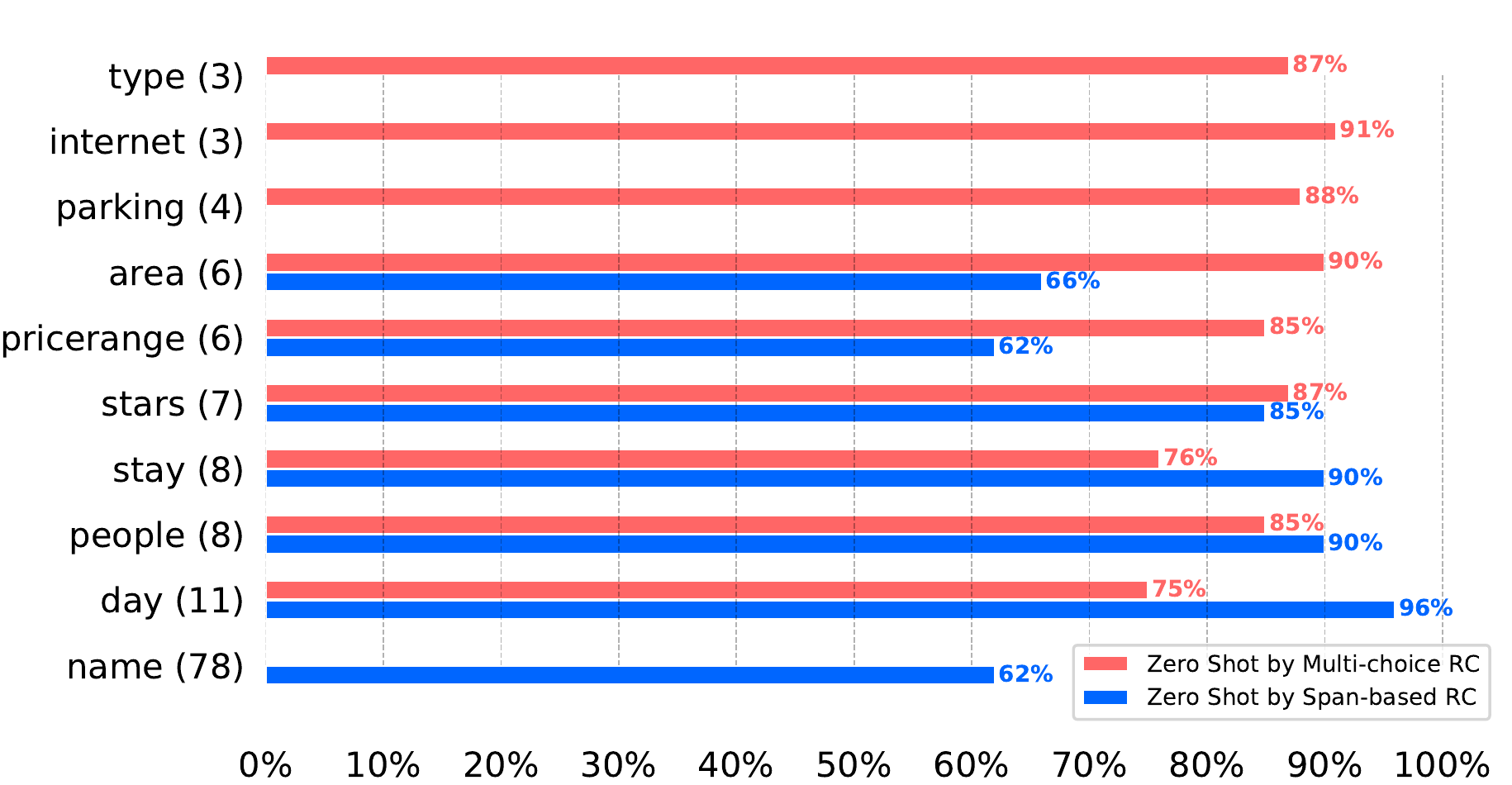}
\caption{Hotel}
\end{subfigure}
\begin{subfigure}[b]{0.5\linewidth}
\centering
\includegraphics[width=1\textwidth]{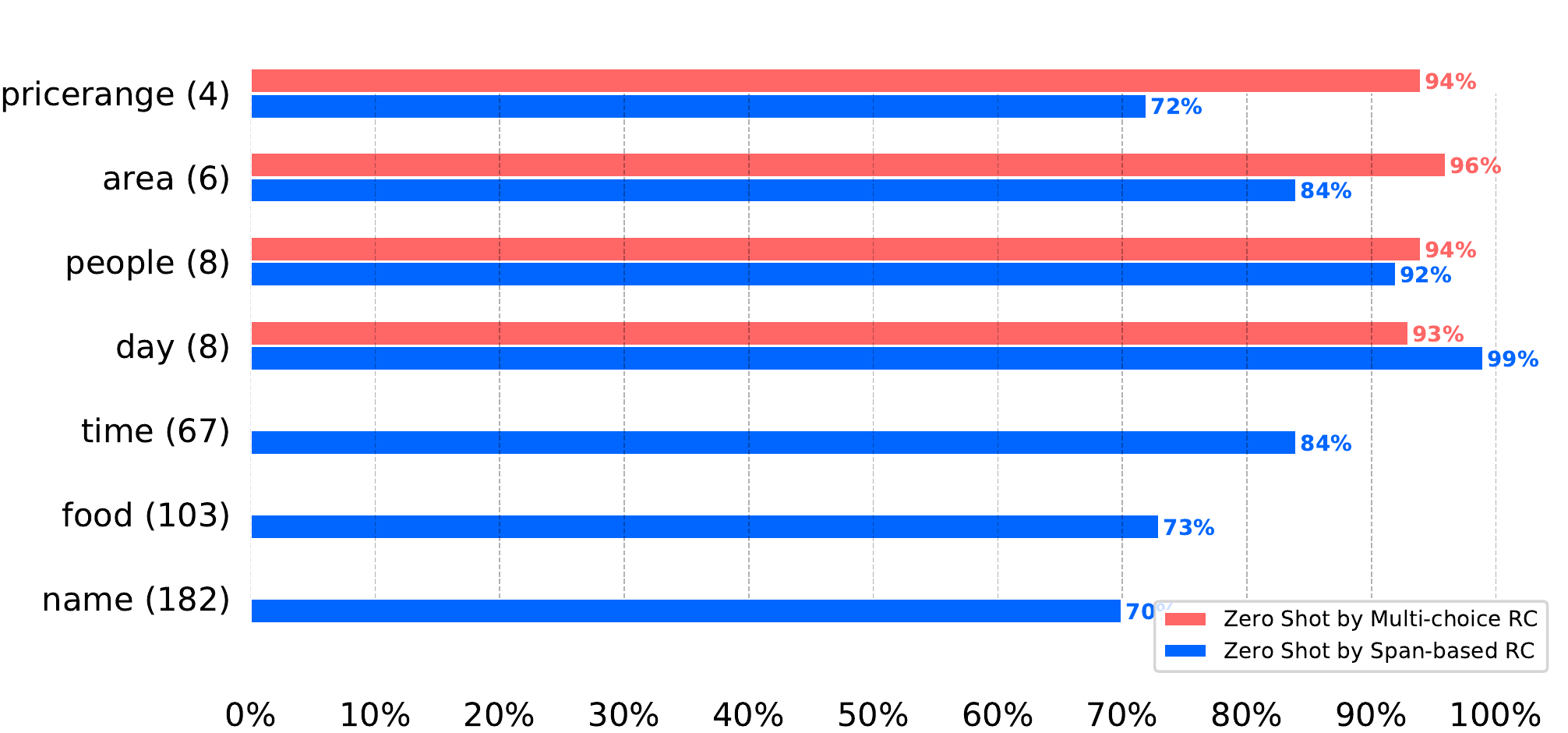}
\caption{Restaurant}
\end{subfigure}
\begin{subfigure}[b]{0.5\linewidth}
\centering
\includegraphics[width=1\textwidth]{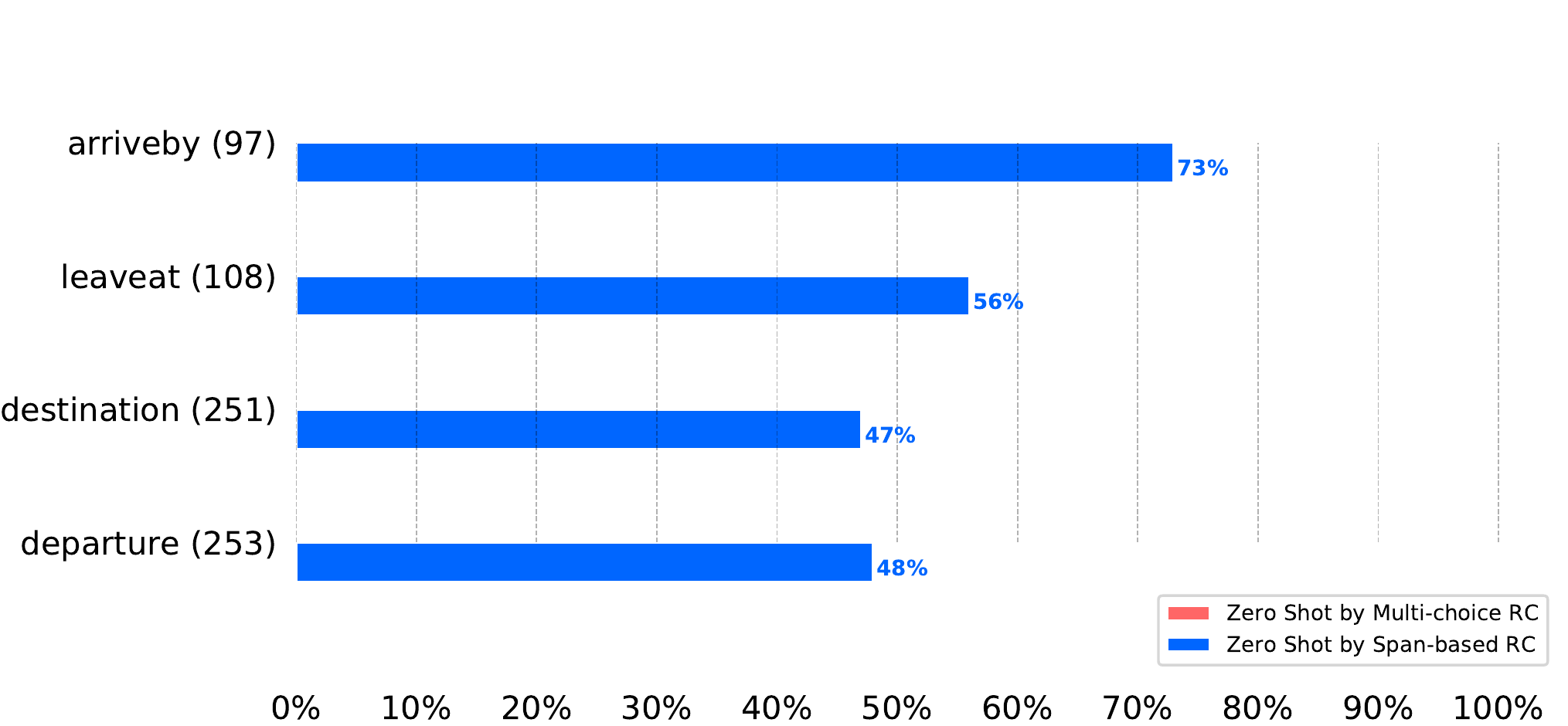}
\caption{Taxi}
\end{subfigure}
\begin{subfigure}[b]{0.5\linewidth}
\centering
\includegraphics[width=1\textwidth]{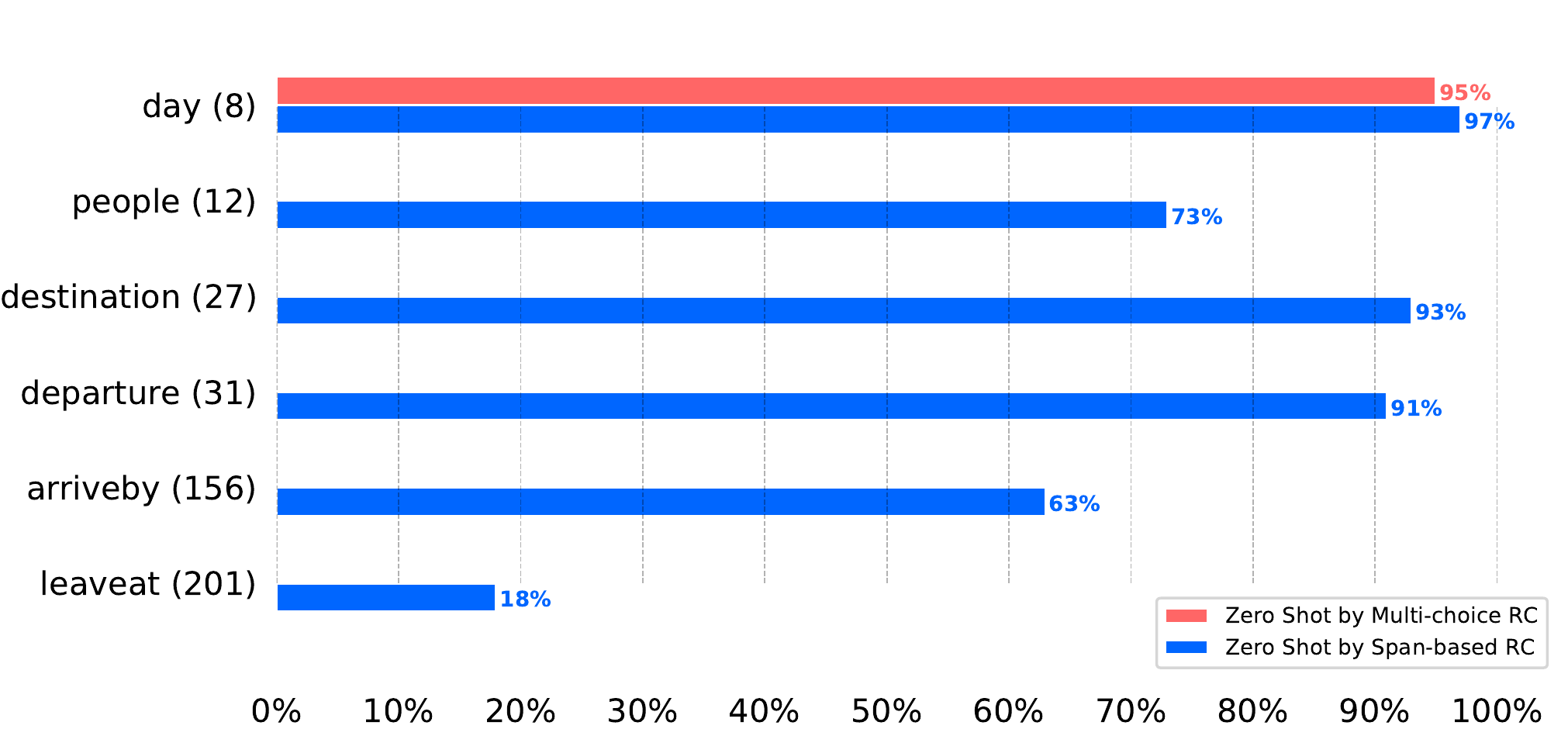}
\caption{Train}
\end{subfigure}
\caption{\small Zero-shot average slot accuracy using multi-choice and span-based RC to DST in hotel, restaurant, taxi, and train domain of MultiWOZ $2.1$\@. The number in parentheses indicates the number of possible values that a slot can take. }
\label{fig:zero_shot}
\end{figure*}

In zero-shot experiments, we want to investigate how would the reading comprehension models behave on MultiWOZ dataset \textit{without} any training on state tracking data. To do so, we train our models on reading comprehension datasets and test on MultiWOZ $2.1$\@. Note that, in this setting, we only take labels in MultiWOZ $2.1$ that are not missing, ignoring the data that is ``None" in the dialogue state. For zero-shot experiments from multiple-choice RC to DST, we take the first fifteen slots in Table~\ref{table:exact_match_rate} that are classified as categorical. For zero shot from span-based RC to DST, we take twenty-seven slots which are extractive expect the first three slots in Table~\ref{table:exact_match_rate}.

Figure~\ref{fig:zero_shot} summarizes the results for hotel, restaurant, taxi and train domain in MultiWOZ 2.1. For attraction domain, please refer to the supplementary section~\ref{sec:sup_zero_shot}. We can see that most of the slots have an average accuracy of at least 50\% or above in both multiple-choice RC and span-based RC approaches, indicating the effectiveness of RC data. For some slots such as \textit{hotel.stay}, \textit{hotel.people}, \textit{hotel.day}, \textit{restaurant.people}, \textit{restaurant.day}, and \textit{train.day}, we are able to achieve very high zero-shot accuracy (greater than 90\%). The zero-shot setting in TRADE~\citep{wu2019transferable}, where the transfer is from the four source domains to the held-out target domain, fails completely on certain slot types like \textit{hotel.name}. In contrast, our zero-shot experiments from RC to DST are able to transfer almost all the slots.

\begin{table*}[t!]
\centering
\small
\begin{tabularx}{1\textwidth}{|X|c|}
\Xhline{1\arrayrulewidth}
                    \textbf{Example (Span-based RC model prediction is bolded)}          &  \textbf{Ground Truth State Value}         \\ \hline
 Dialogue: ``\textit{….A: sure , what area are you thinking of staying, U:  \textbf{i do not have an area preference} but it needs to have free wifi and parking at a moderate price….}" 

Question: ``\textit{which area is the hotel at?}" (hotel.semi.area) &   don't care \\  \hline

 Dialogue: ``\textit{U: i am looking for something fun to do on the \textbf{east side of town} . funky fun house is my favorite place on the east side...}

Question: ``\textit{which area is the restaurant at?}" (restaurant.semi.area) & east \\ \hline
 Dialogue: ``\textit{U: I need 1 that leaves after 13:30 for bishops stortford how about the tr8017 ? A: \textbf{it leaves at 15:29} and arrives at 16:07 in bishops stortford ....}"

Question: ``\textit{what time will the train leave from the departure location?}" (train.semi.leaveat) & 15:29  \\ \hline

 Dialogue: ``\textit{U: hello i want to see some \textbf{authentic architectures} in cambridge!...}" 

Question: ``\textit{what is the type of the attraction?}" (attraction.semi.type)
&  architecture \\ \hline

 Dialogue: ``\textit{...A: can i help you with anything else ? U: i would like to book a taxi \textbf{from the hong house} to the hotel leaving by 10:15...}" 

Question: ``\textit{where does the taxi leave from?}" (taxi.semi.departure) & lan hong house \\

\Xhline{2\arrayrulewidth}

\end{tabularx}
\caption{\small Zero-shot examples to MultiWOZ $2.1$ by span-based reading comprehension model trained on MRQA dataset. The predicted span by the span-based RC model are bolded.}
\label{table:zero_shot_explain}
\end{table*}

Table~\ref{table:zero_shot_explain} illustrates the zero shot examples for span-based RC model. We can see that although the span-based RC model does not directly point to the state value itself, it usually points to a span that \textit{contains} the ground truth state value and the canonicalization procedure then turns the span into the actual slot value. Such predicted spans can be viewed as \textit{evidence} for getting the ground-truth dialogue state, which makes dialogue state tracking more explainable.

\section{Conclusion}
Task-oriented dialogue systems aim to help users to achieve a variety of tasks. It is not unusual to have hundreds of different domains in modern task-oriented virtual assistants. How can we ensure the dialogue system is robust enough to scale to different tasks given limited amount of data? Some approaches focus on domain expansion by training on several source domains and then adapting to the target domain. While such methods can be successful in certain cases, it is hard for them to generalize to other completely different out-of-domain tasks.

Machine reading comprehension provides us a clear and general basis for understanding the context given a wide variety of questions. By formulating the dialogue state tracking as reading comprehension, we can utilize the recent advances in reading comprehension models. More importantly, we can utilize reading comprehension datasets to mitigate some of the resource issues in task-oriented dialogue systems. As a result, we achieve much higher accuracy in dialogue state tracking across different domains given limited amount of data compared to the existing methods. As the variety of tasks and functionalities in a dialogue system continues to grow, general methods for tracking dialogue state across all tasks will become increasingly necessary. We hope that the developments suggested here will help to address this need.



\bibliography{all}

\begin{thebibliography}{28}
\expandafter\ifx\csname natexlab\endcsname\relax\def\natexlab#1{#1}\fi

\bibitem[{Budzianowski et~al.(2018)Budzianowski, Wen, Tseng, Casanueva, Stefan,
  Osman, and Ga{\v{s}}i\'c}]{budzianowski2018multiwoz}
Pawe{\l} Budzianowski, Tsung-Hsien Wen, Bo-Hsiang Tseng, I{\~n}igo Casanueva,
  Ultes Stefan, Ramadan Osman, and Milica Ga{\v{s}}i\'c. 2018.
\newblock Multiwoz - a large-scale multi-domain wizard-of-oz dataset for
  task-oriented dialogue modelling.
\newblock In \emph{Proceedings of the 2018 Conference on Empirical Methods in
  Natural Language Processing (EMNLP)}.

\bibitem[{Chen(2018)}]{chen2018neural}
Danqi Chen. 2018.
\newblock \emph{Neural Reading Comprehension and Beyond}.
\newblock Ph.D. thesis, Stanford University.

\bibitem[{Devlin et~al.(2019)Devlin, Chang, Lee, and
  Toutanova}]{devlin2019bert}
Jacob Devlin, Ming-Wei Chang, Kenton Lee, and Kristina Toutanova. 2019.
\newblock Bert: Pre-training of deep bidirectional transformers for language
  understanding.
\newblock In \emph{Proceedings of the 2019 Conference of the North American
  Chapter of the Association for Computational Linguistics: Human Language
  Technologies, Volume 1 (Long and Short Papers)}, pages 4171--4186.

\bibitem[{Eric et~al.(2019)Eric, Goel, Paul, Sethi, Agarwal, Gao, and
  Hakkani-Tur}]{eric2019multiwoz}
Mihail Eric, Rahul Goel, Shachi Paul, Abhishek Sethi, Sanchit Agarwal, Shuyang
  Gao, and Dilek Hakkani-Tur. 2019.
\newblock Multiwoz 2.1: Multi-domain dialogue state corrections and state
  tracking baselines.
\newblock \emph{arXiv preprint arXiv:1907.01669}.

\bibitem[{Fisch et~al.(2019)Fisch, Talmor, Jia, Seo, Choi, and
  Chen}]{fisch2019mrqa}
Adam Fisch, Alon Talmor, Robin Jia, Minjoon Seo, Eunsol Choi, and Danqi Chen.
  2019.
\newblock Mrqa 2019 shared task: Evaluating generalization in reading
  comprehension.
\newblock \emph{arXiv preprint arXiv:1910.09753}.

\bibitem[{Gao et~al.(2019)Gao, Sethi, Agarwal, Chung, and
  Hakkani-Tur}]{gao2019dialog}
Shuyang Gao, Abhishek Sethi, Sanchit Agarwal, Tagyoung Chung, and Dilek
  Hakkani-Tur. 2019.
\newblock Dialog state tracking: A neural reading comprehension approach.
\newblock \emph{arXiv preprint arXiv:1908.01946}.

\bibitem[{Goel et~al.(2018)Goel, Paul, Chung, Lecomte, Mandal, and
  Hakkani-Tur}]{goel2018flexible}
Rahul Goel, Shachi Paul, Tagyoung Chung, Jeremie Lecomte, Arindam Mandal, and
  Dilek Hakkani-Tur. 2018.
\newblock Flexible and scalable state tracking framework for goal-oriented
  dialogue systems.
\newblock \emph{arXiv preprint arXiv:1811.12891}.

\bibitem[{Goel et~al.(2019)Goel, Paul, and Hakkani-T{\"u}r}]{goel2019hyst}
Rahul Goel, Shachi Paul, and Dilek Hakkani-T{\"u}r. 2019.
\newblock Hyst: A hybrid approach for flexible and accurate dialogue state
  tracking.
\newblock \emph{arXiv preprint arXiv:1907.00883}.

\bibitem[{Jin et~al.(2019)Jin, Gao, Kao, Chung, and Hakkani-tur}]{jin2019mmm}
Di~Jin, Shuyang Gao, Jiun-Yu Kao, Tagyoung Chung, and Dilek Hakkani-tur. 2019.
\newblock Mmm: Multi-stage multi-task learning for multi-choice reading
  comprehension.
\newblock \emph{arXiv preprint arXiv:1910.00458}.

\bibitem[{Lai et~al.(2017)Lai, Xie, Liu, Yang, and Hovy}]{lai2017race}
Guokun Lai, Qizhe Xie, Hanxiao Liu, Yiming Yang, and Eduard Hovy. 2017.
\newblock Race: Large-scale reading comprehension dataset from examinations.
\newblock In \emph{Proceedings of the 2017 Conference on Empirical Methods in
  Natural Language Processing}, pages 785--794.

\bibitem[{Lee et~al.(2019)Lee, Lee, and Kim}]{lee2019sumbt}
Hwaran Lee, Jinsik Lee, and Tae-Yoon Kim. 2019.
\newblock Sumbt: Slot-utterance matching for universal and scalable belief
  tracking.
\newblock \emph{arXiv preprint arXiv:1907.07421}.

\bibitem[{Liu and Lane(2017)}]{liu2017end}
Bing Liu and Ian Lane. 2017.
\newblock An end-to-end trainable neural network model with belief tracking for
  task-oriented dialog.
\newblock \emph{arXiv preprint arXiv:1708.05956}.

\bibitem[{Liu et~al.(2019)Liu, Ott, Goyal, Du, Joshi, Chen, Levy, Lewis,
  Zettlemoyer, and Stoyanov}]{liu2019roberta}
Yinhan Liu, Myle Ott, Naman Goyal, Jingfei Du, Mandar Joshi, Danqi Chen, Omer
  Levy, Mike Lewis, Luke Zettlemoyer, and Veselin Stoyanov. 2019.
\newblock {RoBERTa}: {A} robustly optimized {BERT} pretraining approach.
\newblock \emph{arXiv preprint arXiv:1907.11692}.

\bibitem[{Mrk{\v{s}}i{\'c} et~al.(2016)Mrk{\v{s}}i{\'c}, S{\'e}aghdha, Wen,
  Thomson, and Young}]{mrkvsic2016neural}
Nikola Mrk{\v{s}}i{\'c}, Diarmuid~O S{\'e}aghdha, Tsung-Hsien Wen, Blaise
  Thomson, and Steve Young. 2016.
\newblock Neural belief tracker: Data-driven dialogue state tracking.
\newblock \emph{arXiv preprint arXiv:1606.03777}.

\bibitem[{Nouri and Hosseini-Asl(2018)}]{nouri2018toward}
Elnaz Nouri and Ehsan Hosseini-Asl. 2018.
\newblock Toward scalable neural dialogue state tracking model.
\newblock \emph{arXiv preprint arXiv:1812.00899}.

\bibitem[{Rajpurkar et~al.(2018)Rajpurkar, Jia, and Liang}]{rajpurkar2018know}
Pranav Rajpurkar, Robin Jia, and Percy Liang. 2018.
\newblock Know what you don’t know: Unanswerable questions for squad.
\newblock In \emph{Proceedings of the 56th Annual Meeting of the Association
  for Computational Linguistics (Volume 2: Short Papers)}, pages 784--789.

\bibitem[{Rajpurkar et~al.(2016)Rajpurkar, Zhang, Lopyrev, and
  Liang}]{rajpurkar2016squad}
Pranav Rajpurkar, Jian Zhang, Konstantin Lopyrev, and Percy Liang. 2016.
\newblock Squad: 100,000+ questions for machine comprehension of text.
\newblock In \emph{Proceedings of the 2016 Conference on Empirical Methods in
  Natural Language Processing}, pages 2383--2392.

\bibitem[{Rastogi et~al.(2017)Rastogi, Hakkani-T{\"u}r, and
  Heck}]{rastogi2017scalable}
Abhinav Rastogi, Dilek Hakkani-T{\"u}r, and Larry Heck. 2017.
\newblock Scalable multi-domain dialogue state tracking.
\newblock In \emph{2017 IEEE Automatic Speech Recognition and Understanding
  Workshop (ASRU)}, pages 561--568. IEEE.

\bibitem[{Rastogi et~al.(2019)Rastogi, Zang, Sunkara, Gupta, and
  Khaitan}]{rastogi2019towards}
Abhinav Rastogi, Xiaoxue Zang, Srinivas Sunkara, Raghav Gupta, and Pranav
  Khaitan. 2019.
\newblock Towards scalable multi-domain conversational agents: The
  schema-guided dialogue dataset.
\newblock \emph{arXiv preprint arXiv:1909.05855}.

\bibitem[{Reddy et~al.(2019)Reddy, Chen, and Manning}]{reddy2019coqa}
Siva Reddy, Danqi Chen, and Christopher~D Manning. 2019.
\newblock Coqa: A conversational question answering challenge.
\newblock \emph{Transactions of the Association for Computational Linguistics},
  7:249--266.

\bibitem[{Shah et~al.(2018)Shah, Hakkani-Tur, Liu, and
  Tur}]{shah2018bootstrapping}
Pararth Shah, Dilek Hakkani-Tur, Bing Liu, and Gokhan Tur. 2018.
\newblock Bootstrapping a neural conversational agent with dialogue self-play,
  crowdsourcing and on-line reinforcement learning.
\newblock In \emph{Proceedings of the 2018 Conference of the North American
  Chapter of the Association for Computational Linguistics: Human Language
  Technologies, Volume 3 (Industry Papers)}, pages 41--51.

\bibitem[{Sun et~al.(2019)Sun, Yu, Chen, Yu, Choi, and Cardie}]{sun2019dream}
Kai Sun, Dian Yu, Jianshu Chen, Dong Yu, Yejin Choi, and Claire Cardie. 2019.
\newblock Dream: A challenge data set and models for dialogue-based reading
  comprehension.
\newblock \emph{Transactions of the Association for Computational Linguistics},
  7:217--231.

\bibitem[{Wen et~al.(2017)Wen, Vandyke, Mrk{\v{s}}{\'\i}c, Ga{\v{s}}{\'\i}c,
  Rojas-Barahona, Su, Ultes, and Young}]{wen2017network}
TH~Wen, D~Vandyke, N~Mrk{\v{s}}{\'\i}c, M~Ga{\v{s}}{\'\i}c, LM~Rojas-Barahona,
  PH~Su, S~Ultes, and S~Young. 2017.
\newblock A network-based end-to-end trainable task-oriented dialogue system.
\newblock In \emph{15th Conference of the European Chapter of the Association
  for Computational Linguistics, EACL 2017-Proceedings of Conference},
  volume~1, pages 438--449.

\bibitem[{Wu et~al.(2019)Wu, Madotto, Hosseini-Asl, Xiong, Socher, and
  Fung}]{wu2019transferable}
Chien-Sheng Wu, Andrea Madotto, Ehsan Hosseini-Asl, Caiming Xiong, Richard
  Socher, and Pascale Fung. 2019.
\newblock Transferable multi-domain state generator for task-oriented dialogue
  systems.
\newblock \emph{arXiv preprint arXiv:1905.08743}.

\bibitem[{Xu and Hu(2018)}]{xu2018end}
Puyang Xu and Qi~Hu. 2018.
\newblock An end-to-end approach for handling unknown slot values in dialogue
  state tracking.
\newblock \emph{arXiv preprint arXiv:1805.01555}.

\bibitem[{Zhang et~al.(2019)Zhang, Hashimoto, Wu, Wan, Yu, Socher, and
  Xiong}]{zhang2019find}
Jian-Guo Zhang, Kazuma Hashimoto, Chien-Sheng Wu, Yao Wan, Philip~S Yu, Richard
  Socher, and Caiming Xiong. 2019.
\newblock Find or classify? dual strategy for slot-value predictions on
  multi-domain dialog state tracking.
\newblock \emph{arXiv preprint arXiv:1910.03544}.

\bibitem[{Zhong et~al.(2018)Zhong, Xiong, and Socher}]{zhong2018global}
Victor Zhong, Caiming Xiong, and Richard Socher. 2018.
\newblock Global-locally self-attentive dialogue state tracker.
\newblock \emph{arXiv preprint arXiv:1805.09655}.

\bibitem[{Zhou and Small(2019)}]{zhou2019multi}
Li~Zhou and Kevin Small. 2019.
\newblock Multi-domain dialogue state tracking as dynamic knowledge graph
  enhanced question answering.
\newblock \emph{arXiv preprint arXiv:1911.06192}.

\end{thebibliography}
\bibliographystyle{acl_natbib}
\appendix
\clearpage
\setcounter{figure}{0}  
\section{Zero shot experiments for Attraction domain in MultiWOZ 2.1}
\label{sec:sup_zero_shot}
\begin{figure}[!ht]
\begin{subfigure}[b]{1\linewidth}
\centering
\includegraphics[width=1\textwidth]{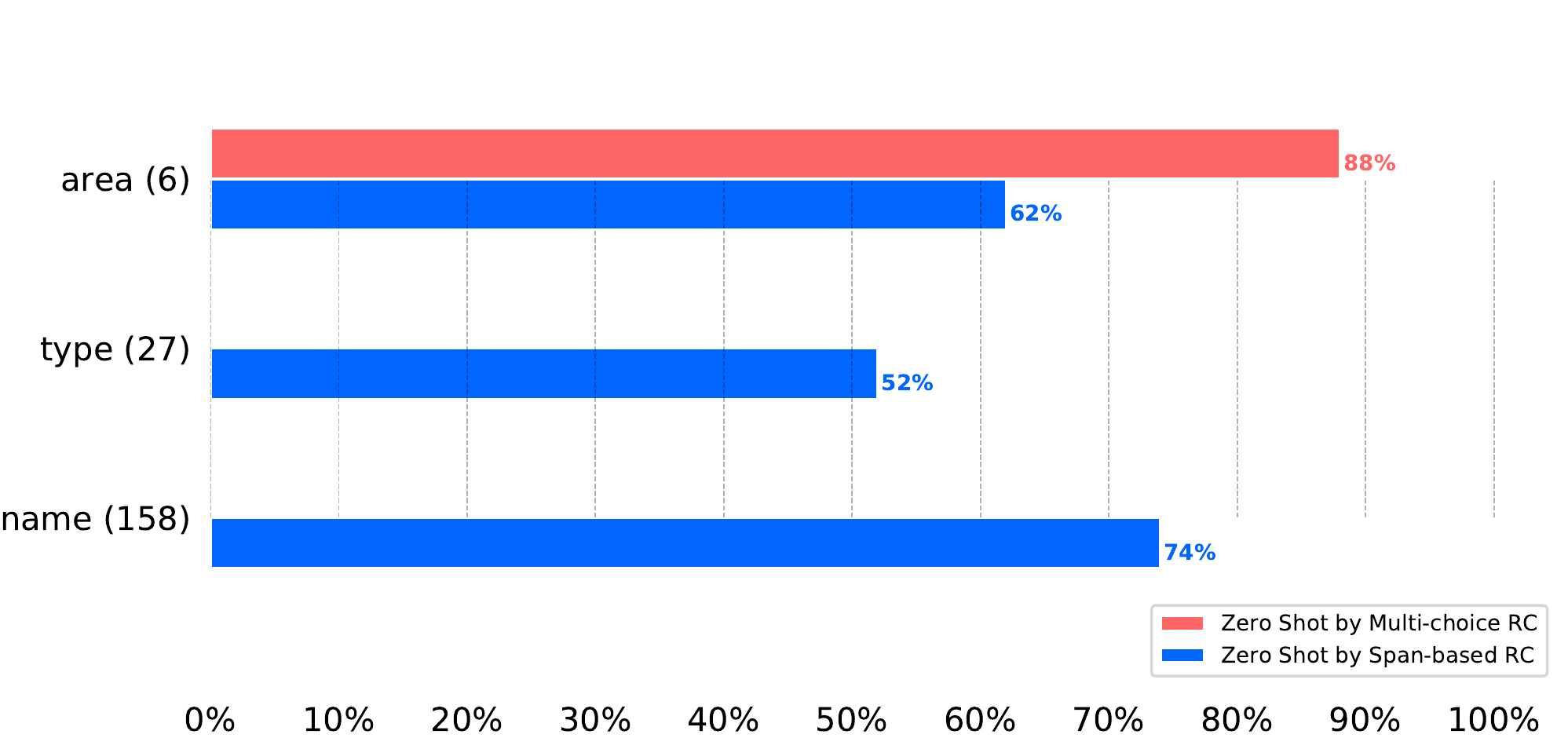}
\caption{Attraction}
\end{subfigure}

\caption{Zero shot average slot accuracy from RC to DST in attraction domain of MultiWOZ 2.1. The number within the brackets associated with each slot name in $y$-axis indicates the number of possible values that a slot can take. }
\label{fig:zero_shot_app}
\end{figure}

\end{document}